\pdfoutput=1
\documentclass[11pt]{article}

\usepackage{acl}

\usepackage{times}
\usepackage{latexsym}
\usepackage{amsfonts}
\usepackage{graphicx}
\usepackage{url} 
\usepackage{amsmath}
\usepackage{ulem}
\usepackage{booktabs}
\usepackage{multirow}

\usepackage[T1]{fontenc}
\usepackage[utf8]{inputenc}

\usepackage{microtype}

\usepackage{listings}

\usepackage{xcolor}

\definecolor{codered}{rgb}{1,0,0}
\definecolor{codegreen}{rgb}{0,0.6,0}
\definecolor{codegray}{rgb}{0.5,0.5,0.5}
\definecolor{codepurple}{rgb}{0.58,0,0.82}
\definecolor{backcolour}{rgb}{0.95,0.95,0.92}

\lstdefinestyle{mystyle}{
  backgroundcolor=\color{backcolour}, commentstyle=\color{codegreen},
  keywordstyle=\color{magenta},
  numberstyle=\tiny\color{codegray},
  stringstyle=\color{codepurple},
  basicstyle=\ttfamily\footnotesize,
  breakatwhitespace=false,
  breaklines=true,
  captionpos=b,
  keepspaces=true,
  numbers=left,
  numbersep=5pt,
  showspaces=false,
  showstringspaces=false,
  showtabs=false,
  tabsize=2
}

\title{Smoothed Contrastive Learning for Unsupervised Sentence Embedding}

\author{
    Xing Wu\textsuperscript{\rm 1,2,3}, Chaochen Gao\textsuperscript{\rm 1,2}\thanks{The first two authors contribute equally.}, Yipeng Su\textsuperscript{\rm 1},  Jizhong Han\textsuperscript{\rm 1},  Zhongyuan Wang\textsuperscript{\rm 3}, Songlin Hu\textsuperscript{\rm 1,2}\thanks{Corresponding author.}
    \\
    \textsuperscript{\rm 1}Institute of Information Engineering, Chinese Academy of Sciences, Beijing, China\\
    \textsuperscript{\rm 2}School of Cyber Security, University of Chinese Academy of Sciences, Beijing, China\\
    \textsuperscript{\rm 3}Kuaishou Technology, Beijing, China
    \\
    \{wuxing,gaochaochen,suyipeng,hanjizhong,husonglin\}@iie.ac.cn
    \\wangzhongyuan@kuaishou.com
}
  
\begin{document}
\maketitle
\begin{abstract}
Unsupervised contrastive sentence embedding models, e.g., unsupervised SimCSE, use the InfoNCE loss function in training. Theoretically, we expect to use larger batches to get more adequate comparisons among samples and avoid overfitting. However, increasing batch size leads to performance degradation when it exceeds a threshold, which is probably due to the introduction of false-negative pairs through statistical observation. To alleviate this problem, we introduce a simple smoothing strategy upon the InfoNCE loss function, termed Gaussian Smoothed InfoNCE (GS-InfoNCE). In other words, we add random Gaussian noise as an extension to the negative pairs without increasing the batch size. Through experiments on the semantic text similarity tasks, though simple, the proposed smoothing strategy brings improvements to unsupervised SimCSE. Our code are available at \href{https://github.com/caskcsg/gsInfoNCE}{https://github.com/caskcsg/gsInfoNCE}.
\end{abstract}

\section{Introduction}

Good sentence representation benefits many natural language processing tasks, and sentence representation learning has been widely studied \cite{logeswaran2018efficient, reimers2019sentence}. 
Contrastive learning has recently been proposed and extensively explored to learn high-quality sentence representations based on the pre-trained language models\cite{devlin2018bert, liu2019roberta}. Contrastive learning aims to learn effective representation by pulling close semantically similar sentences while pushing apart dissimilar ones \cite{hadsell2006dimensionality}. 
Among those unsupervised sentence embedding learning methods with contrastive learning, the latest state-of-the-art method, as far as we know, is unsupervised SimCSE (unsup-SimCSE) \cite{gao2021simcse}. unsup-SimCSE implicitly hypothesizes ``dropout'' as minimal data augmentation and assumes a sentence is semantically more similar to its augmented counterpart than any other sentence. Though simple, unsup-SimCSE works surprisingly well, performing on par with previously supervised counterparts. 
\begin{figure}
\centering
\includegraphics[width=7cm]{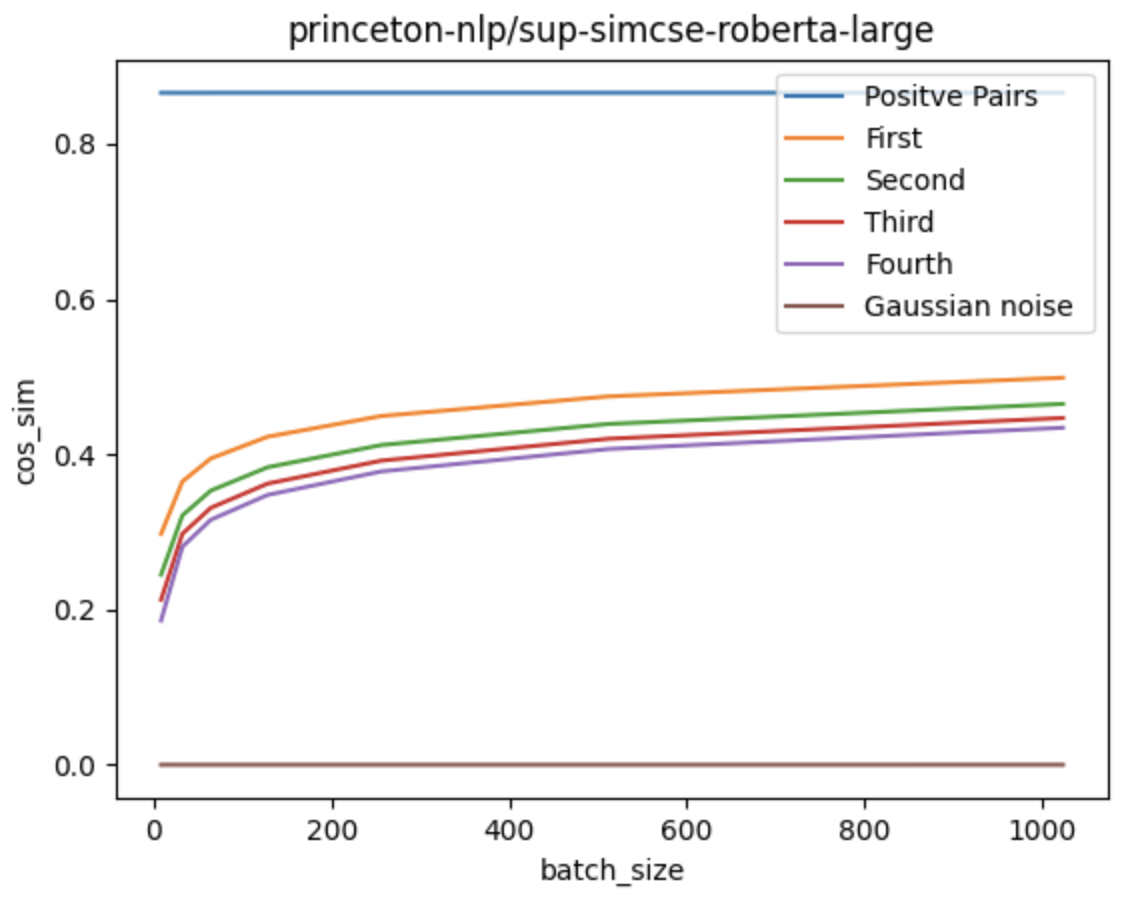}
\caption{The changing trend of the cosine similarities of the negative pairs in the batch. As the batch\_size increases, the mean values of the top 4 cosine similarities also increase, indicating negative pairs with lower confidence exists.}
\label{batch_sim}
\end{figure}

Theoretically, since contrastive learning is carried out among samples within a batch, increasing the batch size will probably bring more adequate comparisons and avoid overfitting. However, according to the original unsup-SimCSE paper \cite{gao2021simcse}, a larger batch size does not always lead to improvements. The performance even decreases when the batch size exceeds a threshold. We assume that as the batch size increases, more similar sentence samples are probably introduced and easily constitute false-negative pairs, which is detrimental to the learning of the model. 
We design a probing statistical experiment for different batch sizes to verify our assumption. 
We use the currently best semantic textual similarity model, i.e., the SimCSE-RoBERTa$_{large}$  \cite{gao2021simcse} to measure the cosine similarity of sentence pairs. Randomly sampling a batch with $N$ sentences, we measure the similarity between all negative pairs within the batch. We calculate the batch's top 4 mean similarity values. We repeat the procedure 100 times and average to eliminate randomness. As shown in Figure\ref{batch_sim}, the top 4 similarity values increase as the batch size increases. 
It means that, in a larger batch, there will be negative pairs comprised of more similar sentences. 
When the batch size does not exceed a threshold, the negative pairs of similar sentences are hard negatives and good for training. 
But when the batch size exceeds, false-negative pairs with higher similarity are introduced, which will mislead the model training. Therefore, achieving sufficient comparison for samples in a ``confident'' (not too large) batch is particularly important.

As shown in the figure \ref{batch_sim}, Gaussian noise is far away from all samples and can constitute a very confident negative pair with any sample within a batch. Therefore, we propose to add random Gaussian noise as an extension to the negative pairs without increasing the batch size \footnote{A contemporaneous work \cite{zhou2022debiased} has also randomly initialized new negatives based on random Gaussian noises to simulate sampling within the whole semantic space, and devise a gradient-based algorithm to optimize the noise-based negatives.}. 
In other words, we introduce a simple smoothing strategy upon the InfoNCE loss function by simply adding a Gaussian noise term to the denominator, termed Gaussian Smoothed InfoNCE (GS-InfoNCE).
From two perspectives, the Gaussian noise term can be understood as a smoothing strategy. Firstly, the number of negative pairs in a given batch is limited and discrete, and these pairs are used to approximate the negative distribution. We can make the distribution smoother by adding random Gaussian noise to extend the negative pairs. Secondly, from the perspective of the loss function, the denominator of GS-InfoNCE's loss introduces an additional penalty term to avoid overfitting. Through experiments on the semantic text similarity (STS) tasks, GS-InfoNCE outperforms the state-of-the-art unsup-SimCSE by an average Spearman correlation of 1.38\%, 0.72\%,  1.17\% and 0.28\% on the base of BERT-base, BERT-large, RoBERTa-base and RoBERTa-large, respectively.

Our contributions can be summarized as follows: we propose GS-InfoNCE for unsup-SimCSE, by introducing a simple smoothing strategy upon the InfoNCE loss function to bring sufficient comparison for samples without increasing the batch size. Our approach can bring improvements to unsup-SimCSE with different model configurations through experiments. 
\section{Background: Contrastive Learning}
Contrastive learning is a discriminative representation learning framework extensively used for unsupervised representation learning. The core idea is to compare a sentence with a semantically similar one (i.e., positive example) and many semantically dissimilar ones (i.e., negative examples). In this way, the semantically similar sentences are closer in the representation space, while the semantically dissimilar ones are farther apart.

\paragraph{InfoNCE}
 \cite{chen2020simple} propose to take a cross-entropy objective with in-batch negatives, namely the InfoNCE objective function. It is a commonly used loss function for contrast learning by pulling similar sentences closer and pushing dissimilar ones apart in the representation space.
Specifically, given a set of sentence pairs: 
$
\mathcal{D}=\left\{\left(x_{i}, x_{i}^{+}\right)\right\}_{i=1}^{m}
$
, where $x_i$ and $x_{i}^{+}$ are the $i$th pair of semantically related sentences. Let $\mathbf{h}_{i}$ and $\mathbf{h}_{i}^{+}$ denote the semantical representations of $x_i$ and $x_i^{+}$, for a mini-batch with $N$ pairs, the training loss for $(x_i , x_i^{+} )$ is:
\begin{equation}
\label{InfoNCE}
\ell_{i}=-\log \frac{e^{\mathrm{sim}\left(\mathbf{h}_{i}, \mathbf{h}_{i}^{+}\right)\tau}}{\sum\limits_{j=1}^{N} e^{\mathrm{sim}\left(\mathbf{h}_{i}, \mathbf{h}_{j}\right) \tau}}
\end{equation}
where $\tau$ is a temperature hyperparameter and $\mathrm{sim}\left(\mathbf{h}_{i}, \mathbf{h}_{i}^{+}\right)$ is the similarity measurement function, which is typically the cosine similarity function.
% as follows.
% \begin{equation}
% \label{similarity}
% \mathrm{sim}\left(\mathbf{h}_{i}, \mathbf{h}_{i}^{+}\right) = \frac{\mathbf{h}_{i}^{\top} \mathbf{h}_{i}^{+}}{\left\|\mathbf{h}_{i}\right\| \cdot\left\|\mathbf{h}_{i}^{+}\right\|}
% \end{equation}

\paragraph{Unsupervised SimCSE}
The idea of unsup-SimCSE is quite simple: each positive pair takes the same sentence as input and utilizes ``dropout'' as minimal data augmentation. In detail, it takes a collection of sentences $\left\{x_{i}\right\}_{i=1}^{m}$ and use $x_{i}^{+}=x_{i}$. It feeds the same input to the encoder twice by applying different dropout masks on fully-connected layers and attention probabilities in the transformer. Through training, positive pair embeddings obtained are similar in the representation space.

\section{Gaussian Smoothed InfoNCE}
We introduce a Gaussian noise term to the InfoNCE loss function, termed Gaussian Smoothed InfoNCE (GS-InfoNCE). Given a Gaussian distribution as follows:
$
G \sim N\left(\mu, \sigma^{2}\right)
$
, whose mean is $\mu$, and the variance is $\sigma^{2}$, we randomly sample $M$ Gaussian noise vectors from it with the same dimensions as the sentence vector. These vectors constitute high confident negative pairs with each sample in the batch to fill and smooth the representation space. Note that these Gaussian noise vectors will not participate in the positive pair constitution. In that way, the loss function of GS-InfoNCE is denoted as follows:
\begin{equation}
\label{GS-InfoNCE}
\ell_{i}=-\log \frac{e^{\mathrm{sim}\left(\mathbf{h}_{i}, \mathbf{h}_{i}^{+}\right) / \tau}}{\sum\limits_{j=1}^{N} e^{\mathrm{sim}\left(\mathbf{h}_{j}, \mathbf{h}_{i}\right) / \tau} + \lambda \cdot \sum\limits_{k=1}^{M} e^{\mathrm{sim}\left(\mathbf{g}_{k}, \mathbf{h}_{i}\right) / \tau}}
\end{equation}
where $\mathbf{g}_{k}$ is a random Gaussian noise vector, $M$ is the number of Gaussian noise vectors involved in the calculation, and $\lambda$ is a balance hyperparameter.

The python implementation of GS-InfoNCE is quite simple, with only three lines of codes based on the original InfoNCE implementation in unsup-SimCSE.

\begin{lstlisting}[language=Python, 
                    caption=Codes in red are regularization modifications to the original InfoNCE loss, 
                    float=*]
    # ... code from origianl unsup-SimCSE above...
    z1, z2 = pooler_output[:,0], pooler_output[:,1]
    cos_sim = cls.sim(z1.unsqueeze(1), z2.unsqueeze(0))
    <@\textcolor{red}{reg\_random = torch.normal(mean, std, size=(reg\_size, hidden\_size)).to(device)}@>
    <@\textcolor{red}{reg\_cos\_sim = cls.sim(z1.unsqueeze(1), reg\_random.unsqueeze(0))}@>
    <@\textcolor{red}{cos\_sim = torch.cat((cos\_sim, reg\_cos\_sim),1).to(device)}@>
    labels = torch.arange(cos_sim.size(0)).<@\textcolor{black}{long()}@>.to(cls.device)
    loss_fct = nn.CrossEntropyLoss()
    # ... code from origianl unsup-SimCSE below...
\end{lstlisting}

\section{Experiments}
We focus on unsup-SimCSE and replace the original InfoNCE objective loss function with GS-InfoNCE. Following \cite{gao2021simcse}, the main goal of sentence embeddings is to cluster semantically similar sentences. For a fair comparison, we conduct our experiments on seven semantic textual similarity (STS) tasks introduced below and take STS results to compare sentence embedding methods. 

\paragraph{Semantic textual similarity tasks} Semantic textual similarity measures the semantic similarity of any two sentences. STS 2012–2016 \citep{agirre2012semeval,agirre2013sem,agirre2014semeval,agirre2015semeval,agirre2016semeval}and STS Benchmark \cite{cer2017semeval} are widely used semantic textual simiarlty benchmark datasets. Following unsup-SimCSE, we use Spearman correlation\footnote{\url{https://en.wikipedia.org/wiki/Spearman\%27s_rank_correlation_coefficient}} to measure the correlation between the ranks of predicted scores and the ground-truth. 
% For a set of size $n$, the n raw scores $X_{i}, Y_{i}$ are converted to its corresponding ranks $\mathrm{rg}_{X_{i}}, \mathrm{rg}_{Y_{i}}$, then the Spearman correlation is defined as follows 
% \begin{equation}
% r_{s}=\frac{\mathrm{cov}\left(\mathrm{rg}_{X}, \mathrm{rg}_{Y}\right)}{\sigma_{\mathrm{rg}_{X}} \sigma_{\mathrm{rg}_{Y}}}
% \end{equation}
% where $\mathrm{cov}\left(\mathrm{rg}_{X}, \mathrm{rg}_{Y}\right)$ is the covariance of the rank variables, $\sigma_{\mathrm{rg}_{X}}$ and $\sigma_{\mathrm{rg}_{Y}}$ are the standard deviations of the rank variables.

\paragraph{Training details} The training details of unsup-SimCSE can be found in \cite{chen2020simple} and github\footnote{\url{https://github.com/princeton-nlp/SimCSE}}. Our experimental settings are consistent with the original method.
For the Gaussian distribution, we empirically use the standard normal distribution, with $\mu=0, \sigma^{2} =1$.
Additionally, we set $\lambda = 1$ and $M = 3 \times batch\_size$ for all experiments.
As illustrated in Figure 1, we have confirmed that increasing the batch size will introduce false-negative pairs with high similarity, so in our experiments, we set the batch size to a moderate size of 64. Following unsup-SimCSE, we conduct experiments on four commonly used models: BERT-base, BERT-large, RoBERTa-base and RoBERTa-large.
\begin{table}[!t]
\centering
\begin{tabular}{lcc}
\hline  
\textbf{Model} & \textbf{SimCSE} & \textbf{+ GS-InfoNCE} \\
\hline
BERT$_{base}$ &  64  &  64 \\
BERT$_{large}$ & 64  &  64 \\
RoBERTa$_{base}$ & 512  &  64 \\
RoBERTa$_{large}$ & 512  &  64\\
\hline
\end{tabular}
\caption{Comparison of $batch\_size$ with or without using GS-InfoNCE in unsup-SimCSE.}
\label{table1}
\end{table}

\begin{table*}[ht]
\centering
\begin{tabular}{lcccccccc}
\hline  
\textbf{Model}& \textbf{STS12} & \textbf{STS13} & \textbf{STS14} & \textbf{SICK15} & \textbf{STS16} & \textbf{STS-B} & \textbf{SICK-R} & \textbf{Avg.}  \\
\hline
%ConSERT-BERT$_{base}$$\clubsuit$ & 64.64 & 78.49 & 69.07 & 79.72 & 75.95 & 73.97 & 67.31 & 72.74  \\
%+ GS-InfoNCE &  &  &  &  &  &  &  &  \\
%ConSERT-BERT$_{large}$$\clubsuit$ & 70.69 & 82.96 & 74.13 & 82.78 & 76.66 & 77.53 & 70.37 & 76.45  \\
%+ GS-InfoNCE &  &  &  &  &  &  &  &  \\
%\hline
SimCSE-BERT$_{base}$$\clubsuit$ & 68.40 & 82.41 & 74.38 & 80.91 & 78.56 & 76.85 & 72.23 & 76.25 \\
+ GS-InfoNCE & \textbf{70.12} & \textbf{82.57} & \textbf{75.21} & \textbf{82.89} & \textbf{80.23} & \textbf{79.70} & \textbf{72.70} & \textbf{77.63} \\
SimCSE-BERT$_{large}$$\clubsuit$ & 70.88 & 84.16 & 76.43 & \textbf{84.50} & 79.76 & 79.26 & \textbf{73.88} & 78.41 \\
+ GS-InfoNCE & \textbf{73.75} & \textbf{85.09} & \textbf{77.35} & 84.44 & \textbf{79.88} & \textbf{79.94} & 73.48 & \textbf{78.96}  \\
\hline
SimCSE-RoBERTa$_{base}$$\clubsuit$ & 70.16 & 81.77 & 73.24 & 81.36 & 80.65 & 80.22 & 68.56 & 76.57 \\
+ GS-InfoNCE & \textbf{71.12} & \textbf{83.24} & \textbf{75.00} & \textbf{82.61} & \textbf{81.36} & \textbf{81.26} & \textbf{69.62} & \textbf{77.74}  \\
\hline
SimCSE-RoBERTa$_{large}$$\clubsuit$ & \textbf{72.86} & 83.99 & 75.62 & \textbf{84.77} & \textbf{81.80} & 81.98 & 71.26 & 78.90 \\
+ GS-InfoNCE & 71.76 & \textbf{84.91} & \textbf{76.79} & 84.35 & 81.74 & \textbf{82.97} & \textbf{71.71} & \textbf{79.21} \\
\hline
\end{tabular}
\caption{Sentence embedding performance on semantic textual similarity (STS) test sets in terms of Spearman’s correlation. $\clubsuit$ : results from the official published model by the unsup-SimCSE.}
\label{table4}
\end{table*}

\paragraph{Main Results}
We list the experimental results in Table \ref{table4}. On the BERT$_{base}$ model, in terms of Spearman correlation, our GS-InfoNCE brings an average increase of 1.38\% over unsup-SimCSE on seven test sets, and the maximum gain on STS-B reach 2.85\%. On the BERT$_{large}$ model, our GS-InfoNCE gives unsup-SimCSE an average improvement of 0.55\% on the 7 test sets, although there is a slight decrease on the SICK15 and SICK-R data sets.
On the RoBERTa$_{base}$ and RoBERTa$_{large}$ models, we have a similar situation, with an average improvement of 1.17\% and 0.31\% on the 7 test sets.

In general, the improvement brought by GS-InfoNCE to unsup-SimCSE is comprehensive. We can fully surpass the previous best model results with the same or smaller batch size in different model configurations, which well demonstrates that our smoothing strategy has played a key role. We believe that a finer search of the parameters can achieve better results and we leave it to our future work.

\paragraph{Analysis: Effect of hyperprarameter $M$}
Gaussian random noise constitutes high-confidence negative pairs with the sentences in a batch. $M$ is the number of Gaussian noise vectors involved in the GS-InfoNCE calculation. We further explore the influence of $M$ on the performance of GS-InfoNCE on BERT$_{base}$. We reuse the hyperparameters of the best-performing model and only vary the hyperparameter $M$. For each $M$, we train the model until convergence and then select the checkpoint that performs the best on the validation set to evaluate on the test set. The performance statistics are listed in Table \ref{M}. As $M$ becomes larger, the performance of GS-InfoNCE on the test set slowly improves. When $M=3$, the best performance is reached, after which the model performance begins to decline. In general, GS-InfoNCE is not sensitive to $M$ (recommend $<8$), making it feasible to apply easily in practical applications. 

% \begin{figure}
% \centering
% \includegraphics[width=8cm]{Trend-M.jpg}
% \caption{The changing trend of model performance when M increases.}
% \label{trend_M}
% \end{figure}

\begin{table}[!b]
\centering
\begin{tabular}{lcccc}
\hline  
\textbf{bs=64} & \textbf{0$\times$}& \textbf{0.5$\times$} & \textbf{1$\times$}& \textbf{2$\times$} \\
\hline
BERT$_{base}$ & 76.25 & 76.96 & 76.90 & 77.11  \\
\hline
\textbf{bs=64} & \textbf{3$\times$}& \textbf{4$\times$} & \textbf{8$\times$}& \textbf{16$\times$}  \\
\hline
BERT$_{base}$ & \textbf{77.63} & 76.81 & 76.94 &  75.57 \\
\hline
\end{tabular}
\caption{Effect of the hyperprarameter $M$ on BERT$_{base}$. We set $M$ as a multiple of batch size (bs=64). \textbf{0$\times$} means the original SimCSE without using GS-InfoNCE.}
\label{M}
\end{table}

\section{Related Work}
 Deep and wide models are prone to overfitting, and thus regularization strategies are important to improve their generalization ability. Among them, smoothing is a very commonly used method.
 \cite{szegedy2016rethinking, muller2019does} propose to use label smoothing as a regularization method that makes the clusters between categories more compact and avoids adversarial examples with over high confidence. Text smoothing\cite{wu2022text, zhu2019soft} also seems to be able to bring further improvements in tasks such as text classification and machine translation by smoothing the one-hot representation of the input text into the probability distribution representation of the dictionary. Our GS-InfoNCE can also be regarded as a smoothing strategy that makes the distribution of negative samples smoother by introducing multiple random Gaussian noise vectors as an extension of the negative examples. Compared with label smoothing and text smoothing, GS-InfoNCE directly uses the standard Gaussian distribution for sampling, largely saving computational costs.

\section{Conclusion and Future Work}
This paper proposes GS-InfoNCE for unsupervised SimCSE methods by introducing a simple smoothing strategy upon the InfoNCE loss function to bring sufficient comparison for samples without increasing the batch size. In the future, we will explore how to improve the generalization capability of GS-InfoNCE and verify its effectiveness on more contrastive learning methods. 

% Entries for the entire Anthology, followed by custom entries
\bibliography{anthology,custom}
\bibliographystyle{acl_natbib}

\end{document}